\documentclass{bmvc2k}

\usepackage{amsmath,amsfonts,bm}

\def\eqref#1{equation~\ref{#1}}

\def\1{\bm{1}}

\DeclareMathAlphabet{\mathsfit}{\encodingdefault}{\sfdefault}{m}{sl}
\SetMathAlphabet{\mathsfit}{bold}{\encodingdefault}{\sfdefault}{bx}{n}

\usepackage{float}
\usepackage{graphicx}
\usepackage{wrapfig}
\usepackage{array}
\usepackage{multirow}
\title{Stream-based Active Learning by Exploiting Temporal Properties in Perception with Temporal Predicted Loss}

\addauthor{Sebastian Schmidt}{sebastian95.schmidt@tum.de}{1,2}
\addauthor{Stephan Günnemann}{s.guennemann@tum.de}{2}

\addinstitution{
 BMW Group\\
 Munich, Germany
}
\addinstitution{
 Technical University of Munich\\
 Data Analytics and \\
 Machine Learning Group,\\
 Munich, Germany
}

\runninghead{Schmidt, Günnemann}{Temporal Predicted Loss}

\def\etal{\emph{et al}\bmvaOneDot}

\begin{document}

\maketitle

\begin{abstract}
Active learning (AL) reduces the amount of labeled data needed to train a machine learning model by intelligently choosing which instances to label.
Classic pool-based AL requires all data to be present in a datacenter, which can be challenging with the increasing amounts of data needed in deep learning.
However, AL on mobile devices and robots, like autonomous cars, can filter the data from perception sensor streams before reaching the datacenter.
We exploited the temporal properties for such image streams in our work and proposed the novel temporal predicted loss (TPL)\footnote{\label{note1} \url{https://www.cs.cit.tum.de/daml/forschung/tpl/}} method.
To evaluate the stream-based setting properly, we introduced the GTA V streets and the A2D2 streets dataset and made both publicly available.
Our experiments showed that our approach significantly improves the diversity of the selection while being an uncertainty-based method.
As pool-based approaches are more common in perception applications, we derived a concept for comparing pool-based and stream-based AL, where TPL outperformed state-of-the-art pool- or stream-based approaches for different models.
TPL demonstrated a gain of $2.5$ precept points (\text{pp}) less required data while being significantly faster than pool-based methods.
\end{abstract}


\section{Introduction}
\label{sec:Intro}
\begin{figure}[tb]
\centering
\begin{minipage}{.5\textwidth}
\centering
 \subfigure[Pool\label{fig:Pool_Scenario}]{%
   \includegraphics[width=0.49\textwidth]{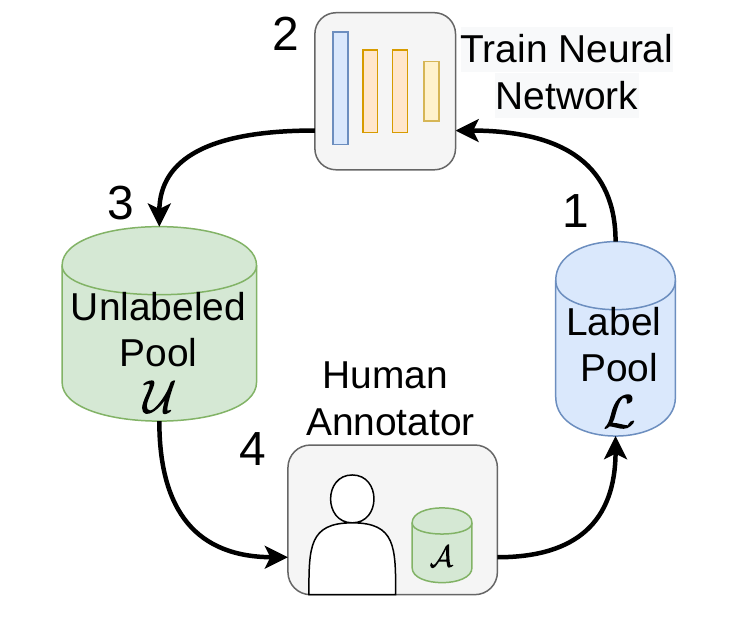}
}%
\subfigure[Stream \label{fig:Stream_Scenario}]{%
   \includegraphics[width=0.49\textwidth]{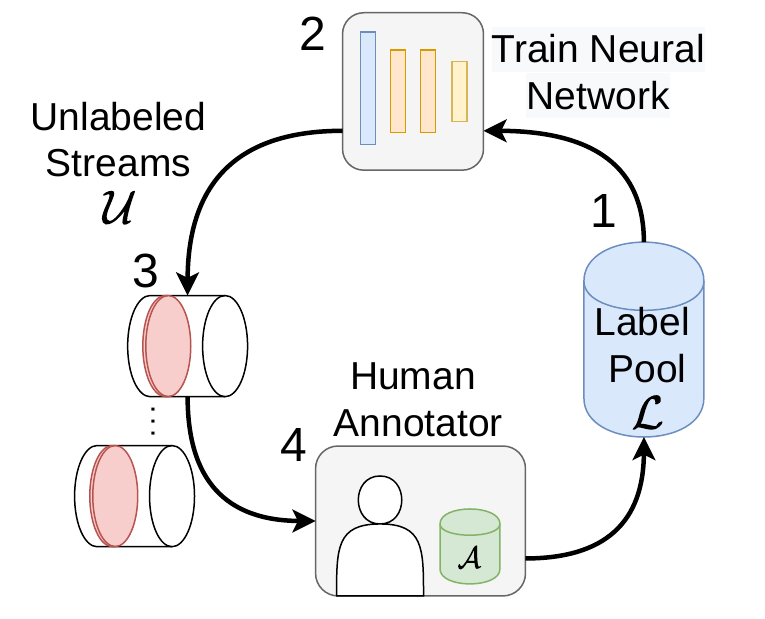}
 }
 \caption{Classic active learning scenarios.}
 \label{fig:Pool_Stream}
\end{minipage}%
\begin{minipage}{.49\textwidth}
\centering
\subfigure[Pool-stream\label{fig:PoolStream_Scenario}]{%
   \includegraphics[width=0.49\textwidth]{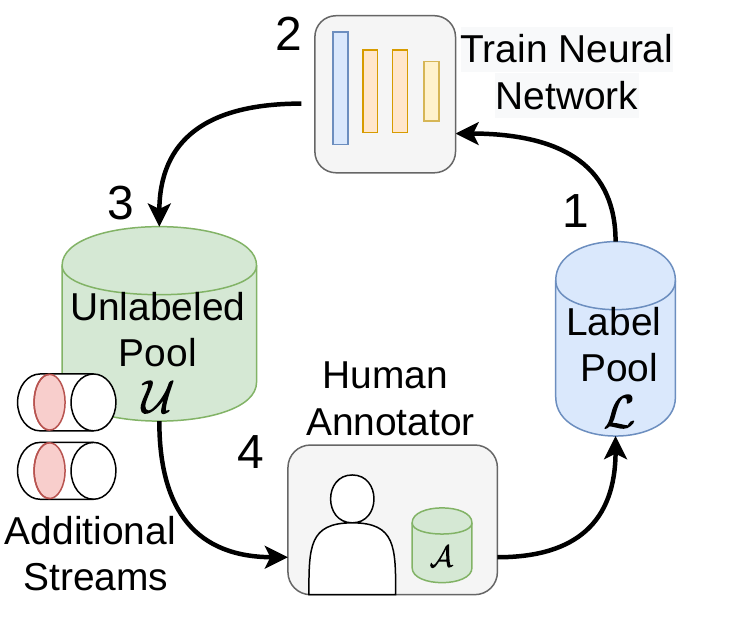}
 }%
\subfigure[Stream-batch\label{fig:StreamBatch_Scenario}]{%
   \includegraphics[width=0.49\textwidth]{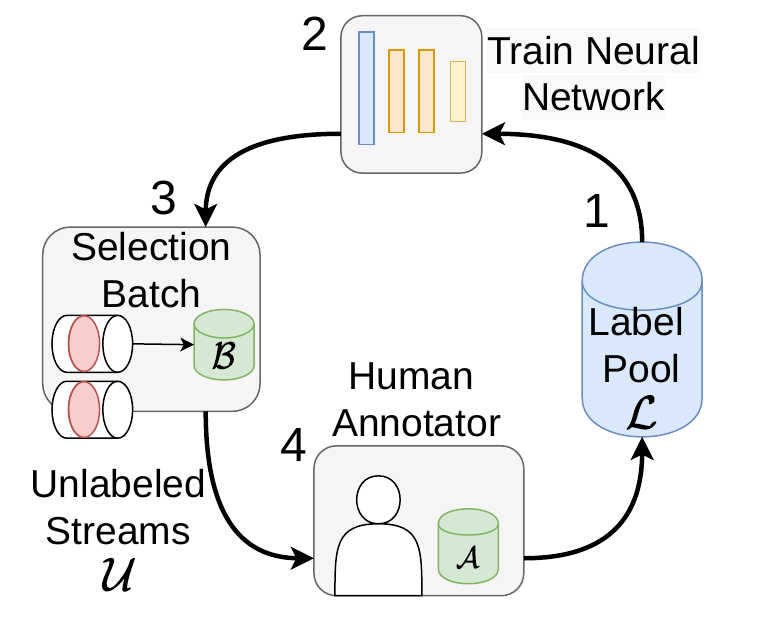}
}
 \caption{Derived active learning scenarios.}
 \label{fig:scenarios}
\end{minipage}
\end{figure}

To apply neural networks in real-world situations, vast amounts of data must be labeled.
Active learning (AL) is a technique to minimize the labeling effort by letting a machine learning model choose the data to be labeled by itself.
The two main scenarios are pool-based and stream-based AL \citep{settles2010}.
The most common pool-based AL scenario, sketched in Figure \ref{fig:Pool_Scenario}, is a cyclic process of selecting batches $\mathcal{B}$ of the most promising samples based on a query function from a pool of data stored on a datacenter.
The model is retrained after the selection to start the next iteration of the AL cycle.
In contrast, stream-based AL, depicted in Figure \ref{fig:Stream_Scenario}, assumes an inflow of samples as a stream and the model decides if a sample should be saved and labeled or disposed.
As the data is examined only once for a decision, the data is not required to be present in a datacenter.
In classic stream-based AL, the model is updated after each selection \citep{settles2010}.
\begin{figure}[htb]
\centering
\includegraphics[width=0.9\textwidth]{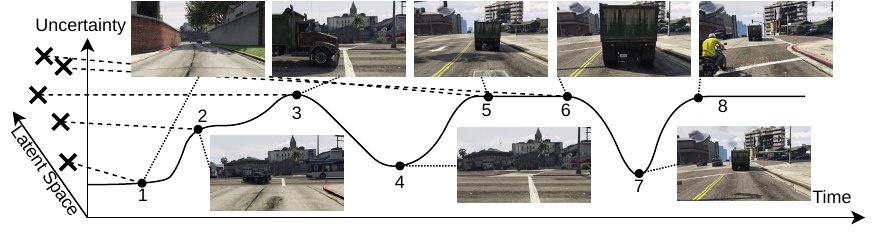}
\caption{GTAVs uncertainty flow - temporal coherent latent space and uncertainty for TPL.}
\label{fig:TPLSketech}
\end{figure}
%
In these scenarios, different selection methods, also called querying strategies, are applied to select the data for labeling.
The three main categories are uncertainty-based, diversity-based and learning-based AL \citep{Ren2022ASO}.
Uncertainty-based AL methods predict or approximate an uncertainty value for each sample \citep{Gal2015,BeluchBcai}.
Diversity-based methods select samples based on the latent space coverage of the dataset. Usually, they perform an optimization, which requires constant access to the labeled and unlabeled dataset.
Learning-based approaches train an additional model to either predict a pseudo-uncertainty value or decide if a sample should be selected directly.
%
The usage of unlabeled data for training the additional model or optimization limits diversity- and learning-based methods primarily to pool-based scenarios.

Real-world perception tasks, like autonomous driving, robotic perception and environmental sensing, are extraordinarily data intensive.
AL is an essential research topic to handle this amount of data for tasks like classification, object detection and semantic segmentation.
However, researchers in this area, particularly autonomous driving, are focused on pool-based AL \citep{Feng2019,Hekimoglu2022,Feng2022}.
Since the data is not generated in the datacenter, data logistics and preparation limit the possibilities to apply and scale AL approaches to open-world perception problems requiring a lot of data.
Divergent from recent efforts, we examine stream-based AL to use these computations' advantages and make AL scaleable for the complexity of real-world applications.
By running stream-based AL directly on mobile devices used in these applications, data can be collected through agents without transferring it to a datacenter.
%

%
To incorporate the real-world challenges, we propose to benchmark AL on camera or sensor stream data directly instead of using the datasets designed for various perception tasks.
We apply AL directly on temporally coherent camera streams to enable agent-wise processing, which reduces preprocessing efforts.
The temporal coherence of the sensor streams enables the representation of the uncertainty as a continuous function of time, depicted in Figure \ref{fig:TPLSketech}. We exploit the temporal coherence and examine the changes in uncertainty to significantly increase the diversity of the selected samples without using optimization techniques requiring access to unlabeled data or extensive optimizations.
%

\textbf{Contributions:} \textbf{(1)} We introduce a new perspective on AL in perception and use it to derive our novel, Temporal Predicted Loss (TPL) AL approach. By incorporating temporal information into uncertainty estimation, TPL increases the diversity of the batch selection. \textbf{(2)} We introduce two operational domain detection datasets, the GTA V streets (GTAVs) based on game recordings and the Audi Autonomous Driving Dataset streets (A2D2s) derived from the A2D2 \citep{Geyer2020} with temporal coherence between the samples. \textbf{(3)} We define a comparison approach between pool-based and stream-based AL and use it to compare our method with state-of-the-art approaches, showing an improvement of $2.5 ~\text{pp}$ less required data.


\section{Related Work}
\label{sec:RelatedWork}
While many authors did significant research on pool-based AL, stream-based AL has lost attention with the rise of deep learning.
%
%
However, stream-based AL is crucial since the amount of vision data that can be uploaded for pool-based AL is limited.
%

\textbf{In pool-based AL} \citet{Sener2018} defined AL as a core-set selection problem, creating a diversity-based AL approach, denoted as CoreSet. A K-center approach selects samples, minimizing the maximum distance to not-selected ones.
Ash \etal \citep{Ash2020} combined the distance representation of the latent space with pseudo labels based on the highest one-hot encoded values to generate gradients to include prediction uncertainty.
Uncertainty-based approaches are mostly sampling-based. Monte Carlo (MC) dropout \citep{Gal2015} uses multiple forward passes with active dropout layers during the prediction to create a distribution over the class predictions, Beluch \etal~\citep{BeluchBcai} used ensembles to create a distribution. The uncertainty is estimated by either the entropy or the mutual information of the distribution. Houlsby \etal~\citep{Houlsby2011} introduced this as Bayesian active learning by disagreement (Bald).
BatchBald \citep{Kirsch2019} takes the diversity of the selected batch into account by calculating the joint mutual information to reduce redundant information in a batch.
%
Loss learning \citep{Yoo2019} is a learning-based approach estimating a pseudo-uncertainty with only one forward pass. A loss learning module is added to specific layers of the prediction network to predict the network's loss, which is used as pseudo-uncertainty for sample selection.
%
Novel learning-based methods like variational adversarial active learning (VAAL) \citep{Sinha2019} use unlabeled data, too. An autoencoder is trained to learn a latent space representation of the data based on the labeled and unlabeled sets. A discriminator model is trained to discriminate between labeled and unlabeled data based on latent space encoding.
Kim \etal~\citep{Kim2021} extended VAAL \citep{Sinha2019} by adding a loss prediction module and included the predicted loss to the latent space.
Caramalau \etal~\citep{Caramalau2021} introduced a more diversity-oriented learning-based approach using unlabeled data with a sequential graph convolutional network (CoreGCN). The authors used the distance between the features of the task model to calculate an adjacency matrix for a graph containing labeled and unlabeled data. A graph neural network predicts the nodes' value for being labeled.

\textbf{Stream-based AL} has a different focus than pool-based AL.
An important topic of research here is conceptual drifts, where the underlying distribution changes over time \citep{Korycki2019,Pham2022}.
In stream-based AL, the selection is often described as a submodular optimization problem where the value of an added labeled sample is dependent on the labels already present. As solving these problems is computationally expensive, stream-based greedy algorithms are an important field of research \citep{Fuji2016}. One approach designed for data summarization is Sieve-Streaming++ \citep{Kazemi2019}.
However, stream-based AL has rarely been used for perception tasks, especially when using deep learning models.
It is more present in low dimensional data streams like chemical plant data \citep{Cacciarelli2022}.
In perception, Mondrian forests have been used in combination with online learning Narr \etal~\citep{Narr}. For non-deep neural network models, online and incremental learning \citep{Chiotellis2018} are often combined with AL.
For neural networks, Puck \etal~\citep{Puck2021} used different uncertainty-based approaches on a mobile robot that stored selected data during operation.
\citet{Senzaki2021} used the Sieve-Streaming++ \citep{Kazemi2019} and applied it to the models' latent space for AL on edge devices (ALED) but neglected temporal and stream properties.
We consider online learning and conceptual drifts as orthogonal problems to this work and focus on connecting pool- and stream-based AL for perception.

\textbf{AL on temporal data} has only been used for pool-based AL in previous works.
%
Bengar \etal~\citep{Bengar2019} used object detection metrics requiring ground truth to build a temporal graph and select samples with energy minimization. For this purpose, the authors created the SYNTHIA-AL dataset with short snippets and a high sampling rate based on the SYNTHIA \citep{Ros2016} dataset.
Schmidt \etal~\citep{Schmidt2020} used the object detection classification uncertainty estimated by the entropy over a time horizon in a pool-based scenario.
Huang \etal~\citep{Huang2018} used temporal information to avoid multiple MC dropout forward passes \citep{Gal2015} for semantic segmentation by combining the uncertainty prediction of a single forward pass with a flow network to calculate the uncertainty as a moving average over a time horizon.

Although many topics have been covered, particularly in pool-based AL, stream-based AL approaches in perception are rare.
The most recent approaches from pool-based AL are diversity-based and learning-based, which are unsuitable for stream-based AL due to the required access to unlabeled data.
Only the uncertainty-based approaches and the loss learning \citep{Yoo2019} approach are applicable for stream-based AL.
Temporal properties have only been used to save the computation of multiple forward passes.
 In stream-based AL, temporal properties remain a relatively unexplored research topic.
In contrast to previous works, we leverage temporal coherence properties in a stream-based AL setup and improve the selection process in terms of speed and performance.
As this research direction is unexplored yet we only consider uncertainty-based query methods as supportive for our work.

\section{Temporal Information in Active Learning}
\label{sec:valueOfTemp}

Most perception datasets do not contain temporal data.
The commonly used datasets Kitti \citep{Geiger2012} and Cityscapes \citep{Cordts2016} aim to have a good diversity to be highly generalizable and do not provide temporally ordered data streams.
The same applies to classification datasets like Cifar10 \citep{cifar}, often used to benchmark AL.
Since these datasets are created for classification, object detection, or semantic segmentation tasks, temporal coherence information is of less importance.
Benchmarking AL on these datasets is sub-optimal and only shows potential label savings on datasets that have been designed for diversity.
Instead, we propose to evaluate AL on camera or sensor streams directly to incorporate real-world challenges like the high correlation between successive samples and avoid additional manual work besides labeling.

Thus, we introduce the A2D2s data and the GTAVs dataset as two benchmark datasets\footnote{\label{note1} \url{https://www.cs.cit.tum.de/daml/forschung/tpl/}}.
These classification datasets feature operational domain detection with temporally coherent frames structured in different recorded drives.
Operational domain detection is a critical task in autonomous driving and mobile robotics to estimate if the current state of the environment is safe to operate.
With the A2D2s we extend the A2D2 \citep{Geyer2020} images by operational domain labels.
To create the GTAVs dataset, we recorded and labeled drives based on the game Grand Theft Auto V (GTA V).
We designed dedicated routes for the individual drives to ensure they overlap free. An overview map is given in Appendix \ref{apx:gtavstreets}.
Each dataset has four different classes describing the environment.
In A2D2 are urban, highway, country road and construction site. For GTAVs, the classes are urban, highway, country road and gravel road. Further details can be found in Appendix \ref{apx:a2d2streets} and \ref{apx:gtavstreets}.
%
%

As elaborated in Section \ref{sec:RelatedWork}, pool-based and stream-based AL are pretty detached.
To enable comparisons of methods for the different scenarios, we modified the original AL cycles from Figure \ref{fig:Pool_Stream} to create the intermediate scenarios pool-stream and stream-batch depicted in Figure~\ref{fig:scenarios}. However, a stream-batch setup was already implicitly used by \citep{Senzaki2021,Puck2021}.
Both scenarios reflect growing datasets closer to mobile robotic and autonomous driving applications.
In detail, they start with a small labeled dataset (1), which is used to train a model (2).
The pool-stream scenario depicted in Figure \ref{fig:PoolStream_Scenario}, reflects a continuous data collection with an unlabeled pool (3) growing at every cycle.
All samples, including the unlabeled ones, can be seen and used multiple times.
To reflect limited recording and transferring capabilities, we change the scenario to a stream-based one.
We remove the pool to create the stream-batch scenario depicted in Figure \ref{fig:StreamBatch_Scenario}, where the current stream is visible for the model only once.
In contrast to the classical stream scenario, a batch $\mathcal{B}$ (3) of a maximum size of $\displaystyle b$ can be selected from a stream in each cycle.

\section{Temporal Predicted Loss - Exploiting Temporal Information in Active Learning}
\label{sec:Methods}
%
For stream-based AL, uncertainty-based approaches remain suitable for choosing a labeling batch $\mathcal{B}$. The methods select the samples with the highest predictive uncertainty estimate $\sigma_p$.
However, these methods cannot avoid overlapping information in the selected batch.

TPL exploits the relation between consecutive uncertainty values to increase the diversity of the selected batch.
By defining the drives of the dataset as consecutively ordered streams, the predictive uncertainty $\sigma_p$ and the stream samples $x$ can be represented as a function of time $\sigma_p(t)$ and $x(t)$ respectively.
Due to the temporal coherence properties of a sensor stream and assuming real-world situations (or their simulations), the change between two samples is continuous.
%
%
%
In situations unknown to the model, the slope of uncertainty exceeds the steady behavior of known situations. By considering the temporal change of the uncertainty, we can detect these situations and filter similar samples with a close temporal relation and similar uncertainty values.
Figure \ref{fig:TPLSketech} depicts an exemplar continuous uncertainty function for a scene flow of GTAVs route 1 and shows the sudden change due to an object appearing (3) and the plateau of closely related scenes (5) and (6).

We use a loss learning module $f_{\sigma}$ \citep{Yoo2019} and its predicted loss as (pseudo-)uncertainty $\tilde{\sigma}_x$ to estimate $\sigma_p$. The sampling of uncertainty-based approaches is problematic for stream-based applications due to increased computational costs. The loss module comprises a global average pool layer, a fully connected layer, and a ReLU activation function. Several modules are attached to specific network layers and concatenated to predict the loss of a given sample $x$. The attachment points are described in Section \ref{sec:Experiments}.
The predicted loss $\tilde{\sigma}_x$ of a sample $x$ is given in Equation \ref{eq:func-t}. 
The loss module and latent space representations depend on the currently selected training set $\mathcal{D}_i$ and are updated after each AL cycle $i$. For readability, we will omit $\mathcal{D}_i$ in further equations.
%
As time is strictly increasing, the time derivative of the predicted loss can be defined by the derivative of the loss prediction module $f_\sigma$ in Equation \ref{eq:lossdt}. \\
\begin{minipage}{0.4\linewidth}
    \vspace{-5pt}
    \begin{equation}
    \label{eq:func-t}
    f_{\sigma}(x(t)|\mathcal{D}_i) = \tilde{\sigma}_{x} \approx \sigma_p.
    \end{equation}
\end{minipage}
\begin{minipage}{0.59\linewidth}
    \begin{equation}
    \label{eq:lossdt}
    \frac{d \tilde{\sigma}_x}{dt} = \frac{d}{dt}f_{\sigma}(x(t)) = \frac{df_\sigma}{dx}\cdot\frac{dx}{dt}
    \centering
    \end{equation}
\end{minipage}
Due to the discrete time steps in data streams, we calculate the derivative of $\frac{d}{dt}\tilde{\sigma}_x$ numerically.
%
We estimate the temporal change of uncertainty via the derivative of the predicted loss $\tilde{\sigma}_x$ and use the temporal predicted loss (TPL) $\frac{d}{dt}\tilde{\sigma}_x$ to quantify the usefulness of a sample x.

%

%
\begin{figure}[tb]
\centering
 \subfigure[Loss learning\label{fig:lloss}]{%
   \includegraphics[width=0.325\textwidth]{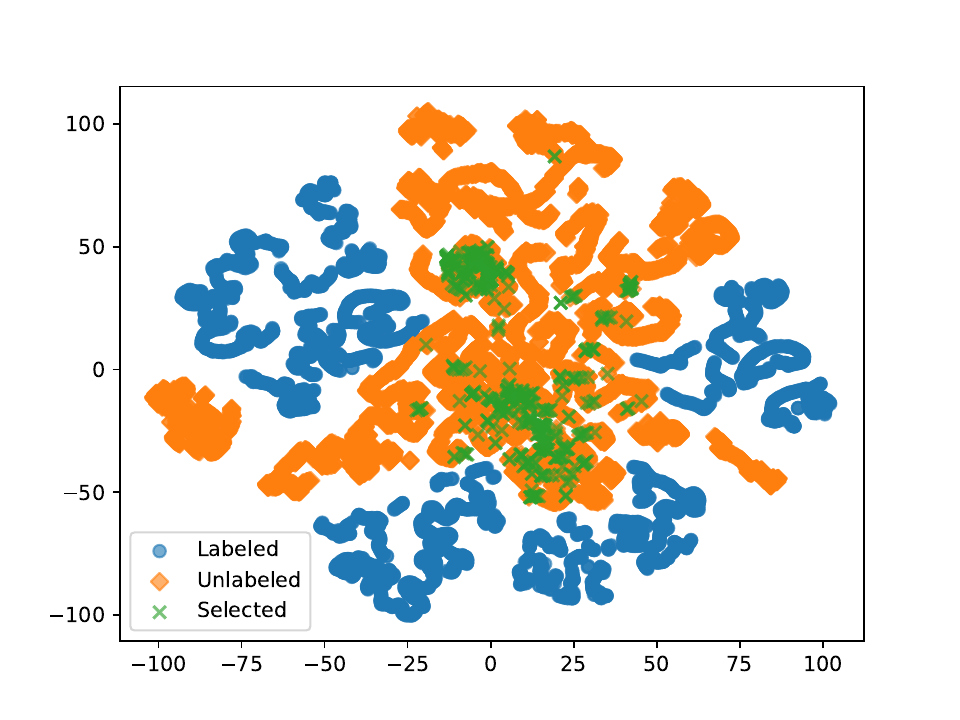}
}%
\subfigure[Temporal predicted loss\label{fig:tpl}]{%
   \includegraphics[width=0.325\textwidth]{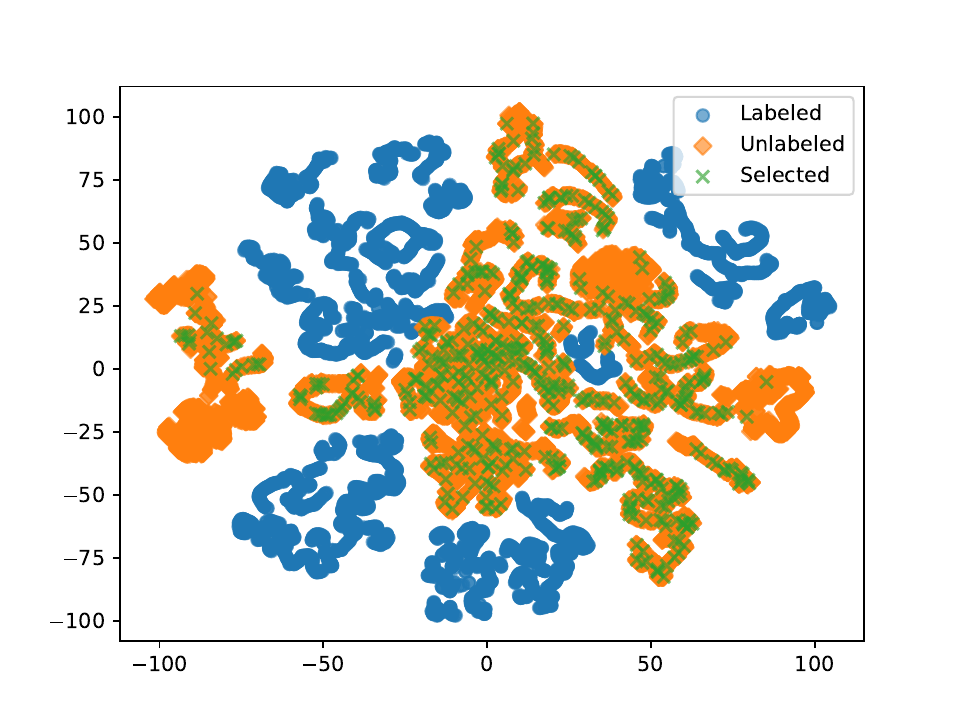}
 }%
 \subfigure[CoreSet \label{fig:cs}]{%
   \includegraphics[width=0.325\textwidth]{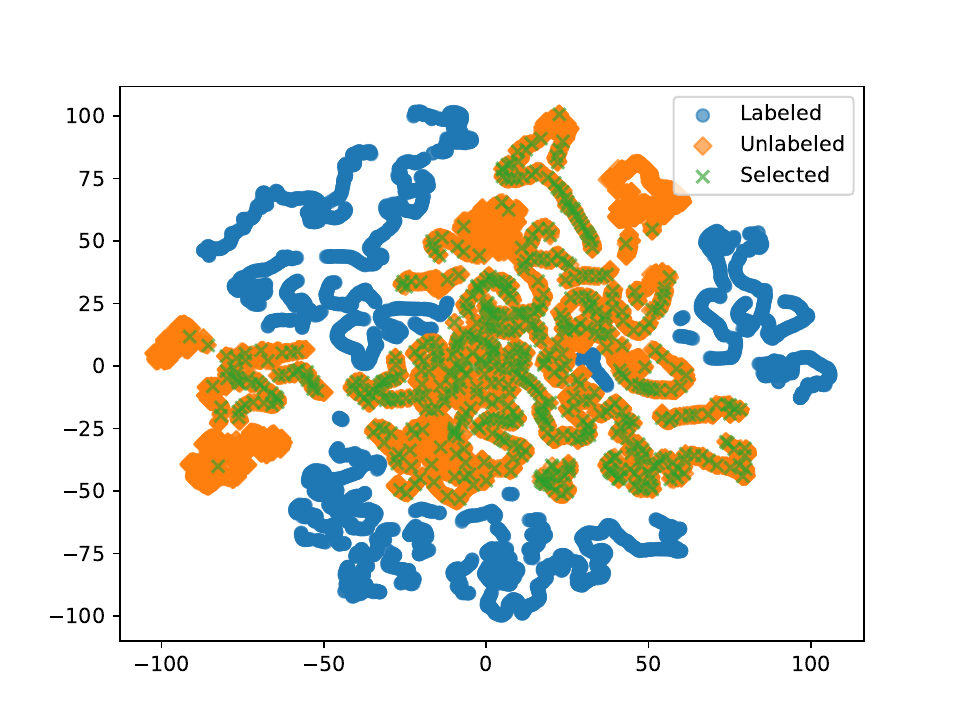}
 }
 \caption{Diversity comparison using t-SNE with a perplexity of 30 for GTAVs route 2. TPL drastically increases the diversity compared to the vanilla approach, such that the diversity is comparable to CoreSet. }
 \label{fig:latentCov}
\end{figure}

Additionally, we modified the loss function of the loss prediction module. The ranking loss function proposed by \citet{Yoo2019} is only capable of learning a relative loss order between two losses $l_i$ and $l_j$ of the task model and the predicted losses $\hat{l}_i$ and $\hat{l}_j$. Therefore, we added an L1-loss with an activation threshold and a scaling factor $\lambda$ to smooth the learning of the loss and reduce the possible offsets between the task loss and its loss prediction.
With the L1 part, we bound the linear offset of the predicted loss, which is impossible when only considering the ranking loss.
The threshold avoids a noisy loss of the module based on the continuously changing loss of the task model.
\begin{equation}
\label{eq:loss_module}
\mathcal{L}_\text{Loss} = \frac{2}{B} \sum\nolimits_{\substack{i=\text{odd}\\j=\text{even}}}^{\text{B}} ~\text{max}(0,\text{Sign}^{*}(l_i-l_j)\cdot (\hat{l}_i-\hat{l}_j)+\xi)
 + \frac{\lambda}{B} \cdot \sum\nolimits_i^B \text{L1}_\text{margin}(l_i,\hat{l_i},\zeta)
\centering
\end{equation}
With $\text{L1}_\text{margin}$ being an L1 loss set to zero if the distance is below a given threshold $\zeta$ and $B$ as batch size. The $\text{Sign}^{*}$ function is modified from the standard sign function, where 0 is included as a negative number and not a positive number, as introduced by \citep{Yoo2019}.
For the final loss, we combined the loss module loss scaled by a factor $\eta$ with a cross-entropy loss, as proposed by \citep{Yoo2019}.
%

\section{Evaluation of Temporal Predicted Loss}
\label{sec:Experiments}

To evaluate our approach, we compare TPL against the vanilla loss learning approach to show the increased diversity in Section \ref{sec:vsvanilla}, state-of-the-art stream-based approaches in Section \ref{sec:streamvsstream} and state-of-the-art pool-based approaches in Section \ref{sec:streamvspool}.

In each AL cycle, a recorded drive is either added to the pool or shown the model as a stream for the pool-stream or stream-batch scenario, respectively. Afterward, the model is trained from scratch and evaluated before a new cycle begins. %
We did not split recordings to avoid overlaps in the train, validation and test split.
Over the cycles, the model can select an additional 50\% of the initial pool size as samples to the labeled pool. To reach this size, we selected 10\% of the current stream for the GTAVs dataset and 20\% for A2D2s. As the length of the recorded streams varies, a different number of samples is selected at each cycle.

The methods are evaluated by their accuracy against the percentage of the labeled data concerning the initial training set and unlabeled pool size. We employ a ResNet18 \citep{He2016} model and a VGG11 \citep{Simonyan15} model with batch normalization, common in AL benchmarks.
To make a fair comparison with sampling methods like BatchBald easier, we extended the classification head of the ResNet18 model to include three fully connected layers with dropout layers in between.
The loss learning modules are attached after each of the four ResNet blocks and four max pooling layers for VGG11.
%
As baselines, we trained the tested models using all the available training data (100\%), a random selection strategy and methods introduced in the related work.
%
%
We conducted a small hyperparameter search to select the appropriate settings to ensure comparability among the different methods.
The supplementary Appendix \ref{ap:experiment} gives detailed information supporting reproducibility.

\subsection{Temporal Coherence Leads to Batch Diversification}
\label{sec:vsvanilla}
In our first experiment, we aim to investigate the impact of temporal relations on the selection of data samples using t-SNE analysis.
%
%
We compare the results of loss learning \cite{Yoo2019}, TPL without the loss function adaption in Equation \ref{eq:loss_module} and the CoreSet \cite{Sener2018} approach in a t-SNE analysis using the GTAVs dataset in Figure \ref{fig:latentCov}.
Since CoreSet is a diversity-based method unsuited for streaming data, we use the pool-stream scenario but the stream-batch for loss learning and TPL.
As shown in Figure \ref{fig:lloss}, the results indicate that loss learning selects a specific cluster of uncertainty with low diversity in the latent space.
In contrast, the TPL method, as shown in Figure \ref{fig:tpl}, demonstrates a high level of diversity, which is slightly less diverse than the diversity-based CoreSet approach shown in Figure \ref{fig:cs}.
%

TPL successfully achieved its objective of enhancing batch diversity without requiring expensive distance calculations. Moreover, as it is a pure uncertainty-based method, it is well-suited for streaming data.
%
%
\begin{figure}[thb]
\centering
\subfigure[GTAVs ResNet18\label{fig:GTA_res18_stream}]{%
   \includegraphics[width=0.48\textwidth]{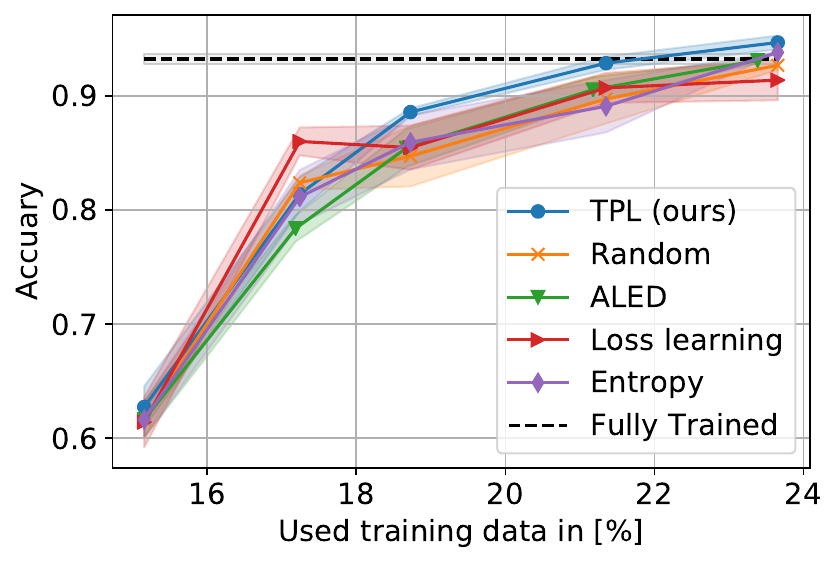}
 }%
 \subfigure[GTAVs VGG11\label{fig:GTA_vgg11_stream}]{%
   \includegraphics[width=0.48\textwidth]{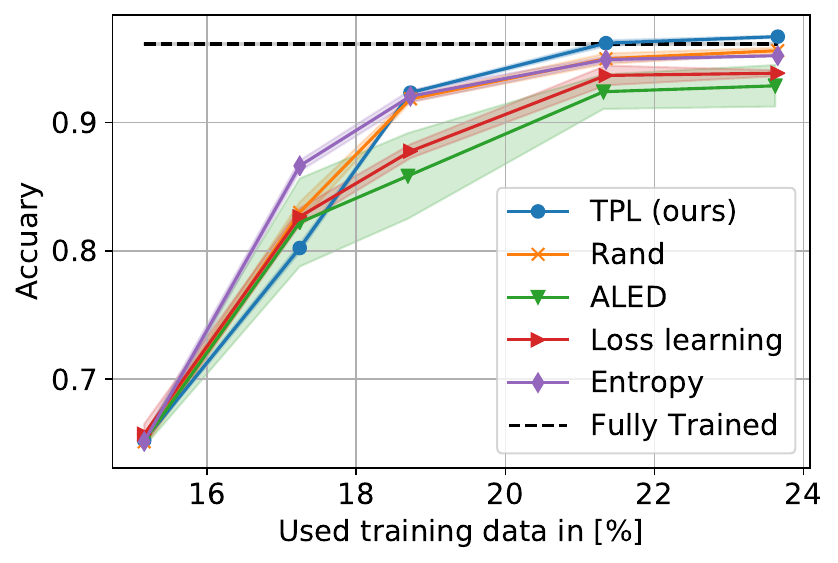}
}%
 \caption{Stream-based comparison - accuracy over used data with indicated standard errors.}
 \label{fig:StreamBasedGTA}
\end{figure}
\subsection{Stream-Based Scenario Evaluation}
\label{sec:streamvsstream}
In these experiments, we compare our TPL method introduced in Section \ref{sec:Methods} with state-of-the-art methods for batch stream-based AL using the stream-batch scenario from Figure \ref{fig:StreamBatch_Scenario}.
We compare TPL for the ResNet18 and VGG11 models on the A2D2s and the GTAVs datasets. As an extension, we used a Deeplabv3 \cite{Chen_2018_ECCV} with ResNet34 backbone to evaluate our method on semantic segmentation labels of A2D2.

\begin{figure}[tb]
\centering
\subfigure[A2D2s VGG11 Stream-based comparison\label{fig:A2D2_vgg11_stream}]{%
   \includegraphics[width=0.48\textwidth]{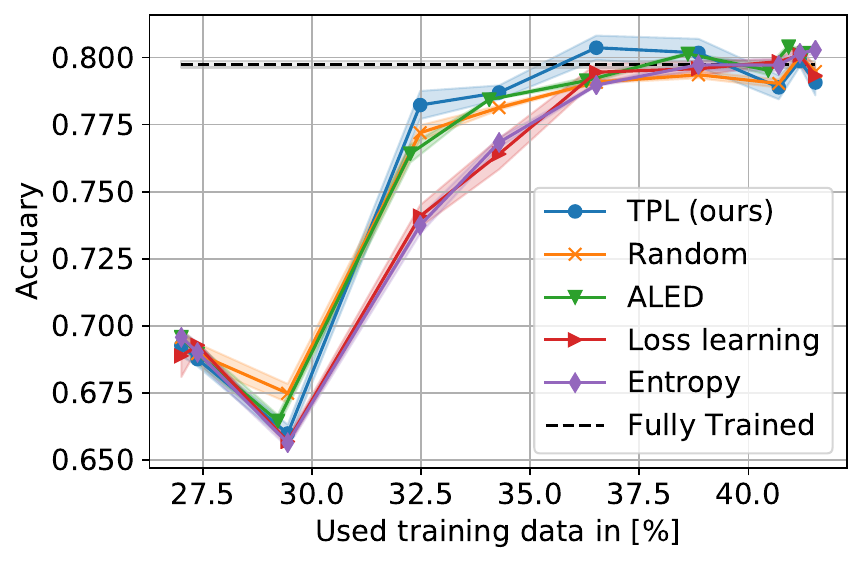}
 }%
 \subfigure[A2D2s VGG11 Pool-based comparison\label{fig:A2D2_VGG11_pool}]{%
   \includegraphics[width=0.48\textwidth]{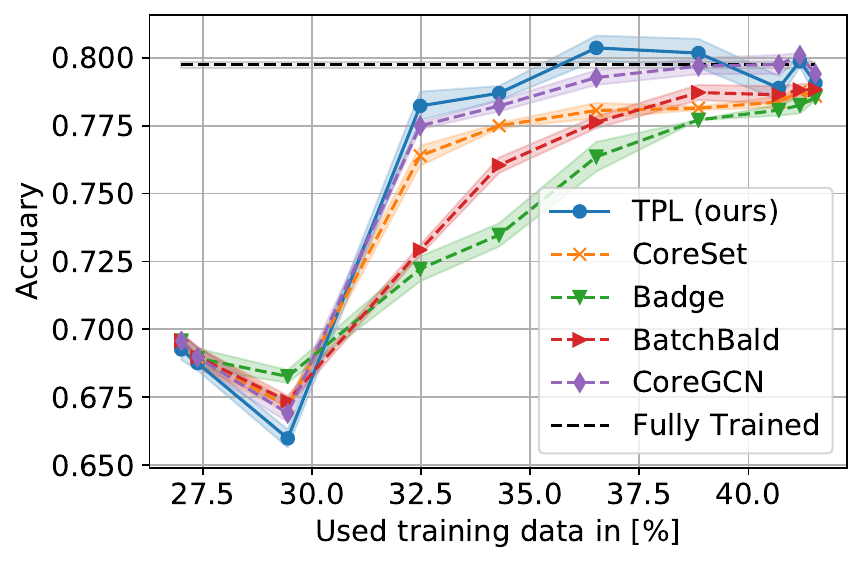}
}%
 \caption{Stream-based and Pool-based comparison for A2D2 - accuracy over used data with indicated standard errors.}
 \label{fig:A2D2_compairson}
\end{figure}

In Figure \ref{fig:StreamBasedGTA} and Figure \ref{fig:A2D2_vgg11_stream}, we compare our TPL with random, MC dropout entropy \cite{Gal2015}, loss learning \cite{Yoo2019} and ALED \cite{Senzaki2021} for A2D2s and GTAVs for ResNet18 and VGG11 with batch normalization.

In Figure \ref{fig:GTA_res18_stream}, we can see that our TPL method intersects the line of the fully trained network at around 21.3\% of the total training data, while the other methods intersect the fully trained network's line between 23\% and 23.8\%. So for ResNet18, TPL achieves the same performance as the network train on all data with only 21\% of the labels, saving up to $2.5 \text{pp}$ more than other AL methods.
For VGG11 in Figure \ref{fig:GTA_vgg11_stream}, TPL intersects the fully trained networks again at 21\% and converges slightly above the performance of the fully trained network.
Interestingly, random is the second-best method for sample selection. A possible explanation is the data's high redundancy and temporal linearity.
The A2D2s dataset, being small and based on real-world data, can result in more significant variations when selecting samples from other recordings. Additionally, the conditions of the stream used for selection may differ from those of the test and validation set (e.g., weather), which can lead to a decrease in performance with new data, as observed in Figure \ref{fig:A2D2_vgg11_stream}. It is possible to exceed the performance of fully-trained network, as the selected subsets may better represent the distribution of the test set than the full dataset.
In Figure \ref{fig:A2D2_vgg11_stream}, TPL achieves the fully trained network's performance at 35.5\% percent of all data. While ALED and random manage a head start before loss learning, the difference gets smaller at 36\% of the data. With ALED intersecting at 37.5\% and loss learning at 38.5\%, TPL saves $2 ~\text{pp}$ more data than state-of-the-art methods.

\begin{wrapfigure}{r}[0pt]{.45\linewidth}
    \begin{center}
   \includegraphics[width=0.45\textwidth]{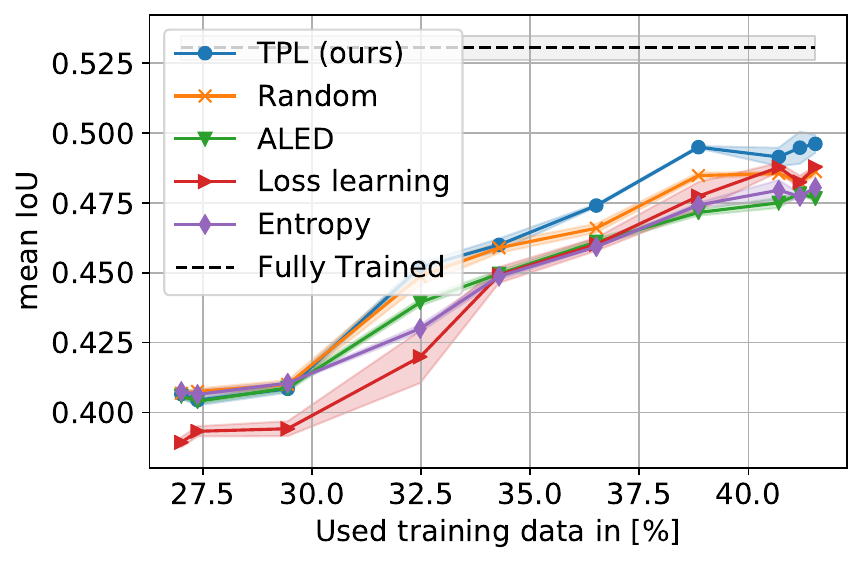}
   \end{center}
   \caption{
         Semantic segmentation comparison with indicated standard errors.
    }
    \label{fig:semSegStream}
\end{wrapfigure}

To underline the flexibility of TPL, we also verify it
on the semantic segmentation dataset part with the DeeplabV3 \citep{Chen_2018_ECCV} model.
The splits, query size and selection parameters are identical to the classification experiments. All details, including loss module attachment points, are given in Appendix \ref{app:semseg}.

Figure \ref{fig:semSegStream} shows the limitation of stream-based AL created by the data logistic advantage. While the pool-based selection process can be continued until a desired performance is reached, the number of cycles for stream-based scenario is tied to the number of streams. Here, all methods fail to reach the fully trained network's performance. Selecting 20\% of each stream is too little for semantic segmentation. Notably, TPL achieves the highest mean Intersection over Union (mIoU) by using only 39\% of the data. The mIoU value is 1\% higher than the second-best methods without any parameter or adaptation. This proves the potential and flexibility to apply TPL to other tasks.
\subsection{Comparing TPL with Pool-Based Scenarios}
\label{sec:streamvspool}
To introduce stream-based strategies as a genuine alternative to pool-based AL, we compare our TPL method with pool-based AL approaches. We use the stream-batch scenario for TPL and the pool-stream scenario for BatchBald \cite{Kirsch2019}, CoreSet \cite{Sener2018}, Badge \cite{Ash2020} and CoreGCN \cite{Caramalau2021}.
To the best of our knowledge, we are the first to compare the different scenarios explicitly.

\begin{figure*}[ht]
\centering
 \subfigure[GTAVs ResNet18\label{fig:GTA_Res18_pool}]{%
   \includegraphics[width=0.48\textwidth]{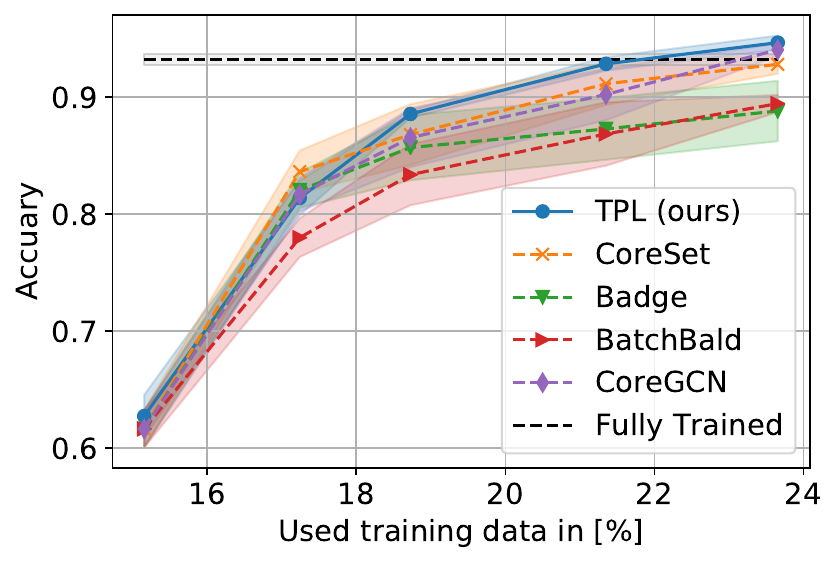}
 }%
 \subfigure[GTAVs VGG11\label{fig:GTA_VGG11_pool}]{%
   \includegraphics[width=0.48\textwidth]{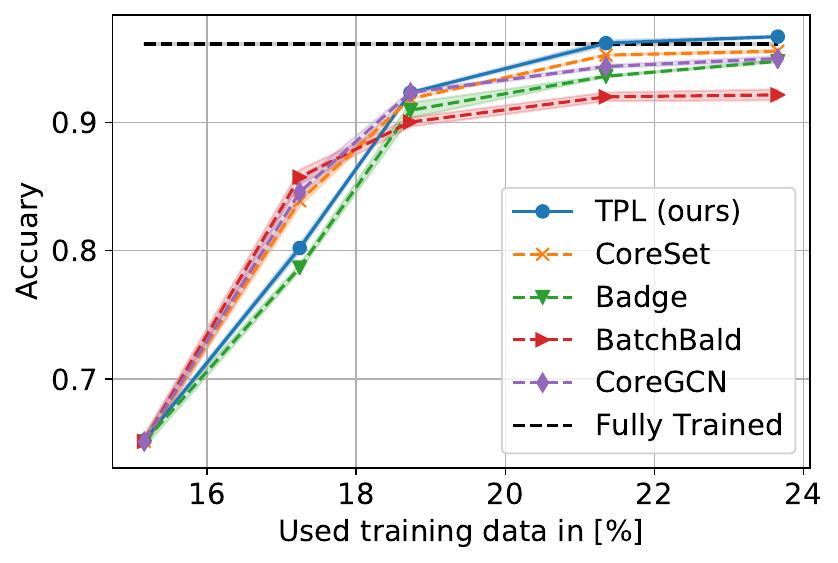}
}%
 \caption{Pool-based comparison - accuracy over used data with indicated standard errors.}
\label{fig:SPcomp}
\end{figure*}

TPL outperforms the second-best method CoreGCN by reaching the fully trained network line with $3~\text{pp}$ fewer data for A2D2 with VGG and $2~\text{pp}$ fewer data for GTAVs with Resnet18 depicted in Figure \ref{fig:A2D2_VGG11_pool} and Figure \ref{fig:GTA_Res18_pool}. However, the pool-based methods do not suffer from the limited selection of the last streams of the A2D2s. For GTAVs with VGG11 illustrated in Figure \ref{fig:GTA_VGG11_pool}, TPL is the only approach crossing the fully trained line, while CoreGCN stays slightly behind. CoreSet and the learning-based CoreGCN considering diversity outperform the other pool-based methods.
Although the pool-based scenario provides an information advantage, pool-based methods without diversity consideration show poor performance.
Our results suggest that diversity is crucial in sample selection on highly redundant datasets.
%
%
%
%

%
%

%
Additionally, we compared the selection times of the different methods on a dedicated machine. Table \ref{tab:selectionTime} shows our computational efficiency, especially against pool-based methods with at least four times.
TPL offers data savings like state-of-the-art pool-based methods while offering the reduced data logistics of stream-based approaches with an at least seven times faster selection.
As most perception problems start from sensor streams (e.g., camera), our TPL method can be used on the mobile device directly for many applications like robotics, autonomous driving and environmental surveillance, saving additional computational costs and data-prepossessing efforts.
%

\begin{table}[ht]
\resizebox{\textwidth}{!}{%
\begin{tabular}{lcccccccc}
Method & Loss learn. & TPL & Entropy    & ALED  & BatchBald & Badge     & CoreSet  & CoreGCN  \\
\hline
Time [s]  & \textbf{4.5}  & \underline{4.6}  & 6.3 & 427.2 & 835.2 & 49.7 & 32.7 & 49.7 \\
\end{tabular}%
}
\caption{Mean selection time on GTAVs/ResNet18 - Nvidia A6000, Intel Xeon Gold 6248R.}
\label{tab:selectionTime}
\centering
\end{table}

\section{Conclusion and Future Work}
Our work explored temporal coherence in data for stream-based active learning (AL) in perception.
We proposed a novel approach leveraging temporal coherence to increase the diversity of uncertainty-based AL methods. We combined it with a modification of the loss learning module \citep{Yoo2019} in our temporal predictive loss (TPL) method.
To reflect the properties of stream-based data collection, we introduced the A2D2 streets (A2D2s) and GTA V street (GTAVs) datasets and made them publicly available.
Additionally, we introduced the pool-stream scenario to enable comparisons with the stream-batch scenario.

Our experiments showed that TPL significantly increases the diversity of the selection compared to vanilla loss learning and achieves almost the diversity of diversity-based methods.
Compared to state-of-the-art stream-based AL methods, TPL achieves the highest accuracy with $2.5 ~\text{pp}$ less annotated data, using only 21\% of the GTAVs dataset.
To the best of our knowledge, our work provided the first comparison between stream-based and pool-based AL using pool-stream and stream-batch scenarios, showing that our TPL method outperforms pool-based methods while being seven times faster.
Overall, this study represents an essential step towards stream-based AL for perception tasks, demonstrating the effectiveness of stream-based approaches. Furthermore, it highlights stream-based approaches as a viable alternative to pool-based approaches.
The additional advantage in data logistics of stream-based AL is a significant consideration for enabling large-scale AL.
%

%
In future work, we plan to investigate alternative methods for uncertainty estimation.
One potential avenue of research is integrating uncertainty estimation with diversity measurements, which have shown promising results in pool-based approaches.
Additionally, we plan to extend our approach to exploit temporally coherent vision data in lane and object detection.
Finally, a dedicated dataset for stream-based AL in mobile robot perception would further the field and provide valuable resources to the research community.

\section*{Acknowledgement}
I want to thank my colleagues from BMW and TUM for the support and helpful discussions during this project.

\bibliography{bmvc_final}


\newpage
\appendix

\section{The A2D2 Streets Dataset}
\label{apx:a2d2streets}
\begin{figure}[htb]
    \begin{center}
   \includegraphics[width=0.6\textwidth]{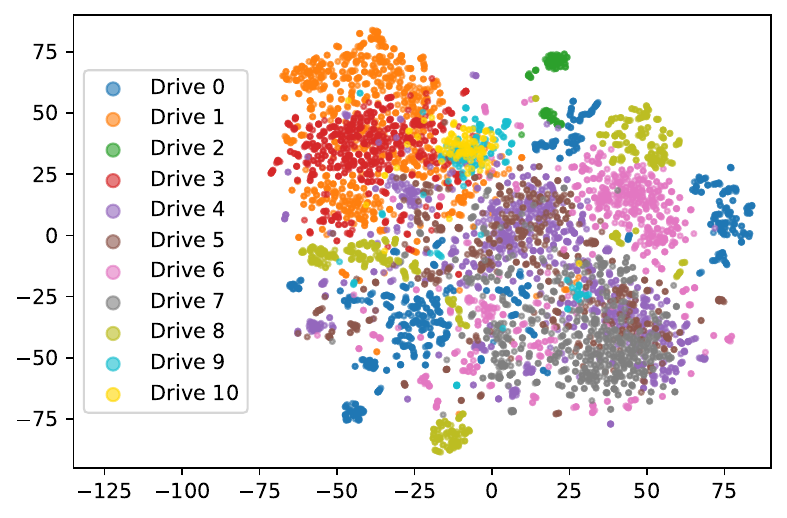}
   \end{center}
   \caption{
         t-SNE analysis with perplexity 30 of the different recorded drives from the training set in A2D2s.
    }
    \label{fig:T-SNE}
\end{figure}

We created a scene classification dataset for an operational domain detection task. This task is essential for autonomous vehicles as it reflects whether they can operate safely in this environment. For example, construction sites are a domain where more caution is required.
For the dataset, we used the image data of the A2D2 
\citep{Geyer2020} which provides temporally coherent frames structured in different drives.
We assigned the classification labels urban, highway, country road and construction site to the images describing the current driving environment.
The dataset contains several recorded drives in southern Germany, with around 680 frames on average per recording.
The frames are timestamped with a high frequency of up to 10 Hz so that the temporal change of the samples can be evaluated meaningfully.
Although the rate is not constant due to sensor synchronization, the optical flow remains stable and does not get lost.
The temporal coherence with a high frequency of sampled images brings the risk of selecting redundant samples in a batch.
Due to the nature of the drives, the latent space representation is naturally clustered in the specific drives shown in Figure \ref{fig:T-SNE}.
\begin{figure}[htb]
\centering
 \subfigure[Construction site]{%
   \includegraphics[width=0.24\textwidth]{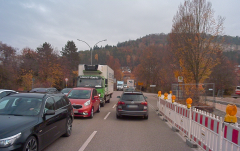}
 }%
\subfigure[Country road]{%
\includegraphics[width=0.24\textwidth]{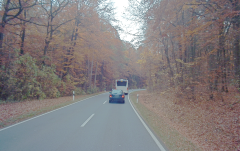}
}%
\subfigure[Highway]{%
\includegraphics[width=0.24\textwidth]{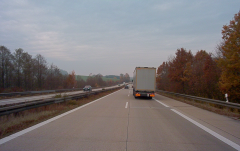}
}%
\subfigure[Urban]{%
\includegraphics[width=0.24\textwidth]{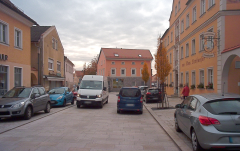}
}
 \caption{Overview of the different classes in the A2D2 streets dataset.}
\label{fig:A2D2Examples}
 \end{figure}
The recorded drives are not split and divided as a whole recording into an initial labeled pool and unlabeled pool for training as well as validation and test set as shown in Table
\ref{tab:A2D2}. In the stream-based setups, the unlabeled drives are fed as streams into the AL algorithm.
The images have been resized to $120 \times 72$ pixels.
The training dataset has been shuffled.
\begin{table}
\centering
 \begin{tabular}{m{0.2\textwidth} m{0.65\textwidth}}
  \multicolumn{1}{c}{Assignment} & \multicolumn{1}{c}{Sessions}\\
  \hline \\

  initial labeled & \text{20181107\_132730} ~\text{20181108\_091945} \\
    & \text{20181107\_133258} ~\text{20181108\_084007} ~\text{20180807\_145028} \\
   unlabeled pool & \text{20180810\_142822} ~\text{20180925\_135056} ~\text{20181008\_095521} \\
   & \text{20181107\_132300} ~\text{20181204\_154421} ~\text{20181204\_170238} \\

  validation set  & \text{20180925\_101535} ~\text{20181016\_125231} ~\text{20181204\_135952}\\
  test set & \text{20180925\_124435} ~\text{20181108\_123750} ~\text{20181108\_103155}\\
 \end{tabular}
\caption{This table shows the dataset split into internal labeled and unlabeled pool training set as well as validation and test set for A2D2s. }
\label{tab:A2D2}
 \end{table}

\section{The GTA V Streets Dataset}
\label{apx:gtavstreets}
We created the GTA V streets dataset \footnote{\label{note2}\url{https://www.cs.cit.tum.de/daml/forschung/tpl/}} as the first classification dataset designed for active learning, having temporal coherence. Like the A2D2 streets dataset, we designed an operation domain detection task for our dataset and added the labels for highway, urban, country road and gravel road. However, the dataset was recorded from a game, not the real world like A2D2, so the environment is more manageable and potential variables like weather can be avoided.

\begin{figure}[htb]
\centering
 \subfigure[Gravel Road]{%
   \includegraphics[width=0.24\textwidth]{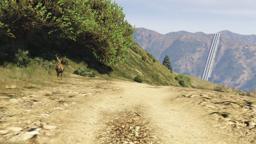}
 }%
  \subfigure[Country road]{%
   \includegraphics[width=0.24\textwidth]{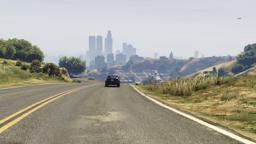}
 }%
  \subfigure[Highway]{%
   \includegraphics[width=0.24\textwidth]{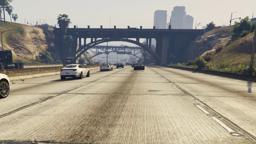}
 }%
  \subfigure[Urban]{%
   \includegraphics[width=0.24\textwidth]{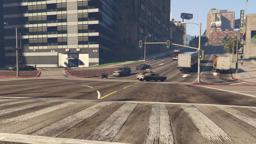}
 }%
 \caption{Overview of the different classes in the GTAV streets dataset.}
\label{fig:GTAExamples}
 \end{figure}
\begin{figure}[ht]
\centering
\includegraphics[width=0.72\textwidth]{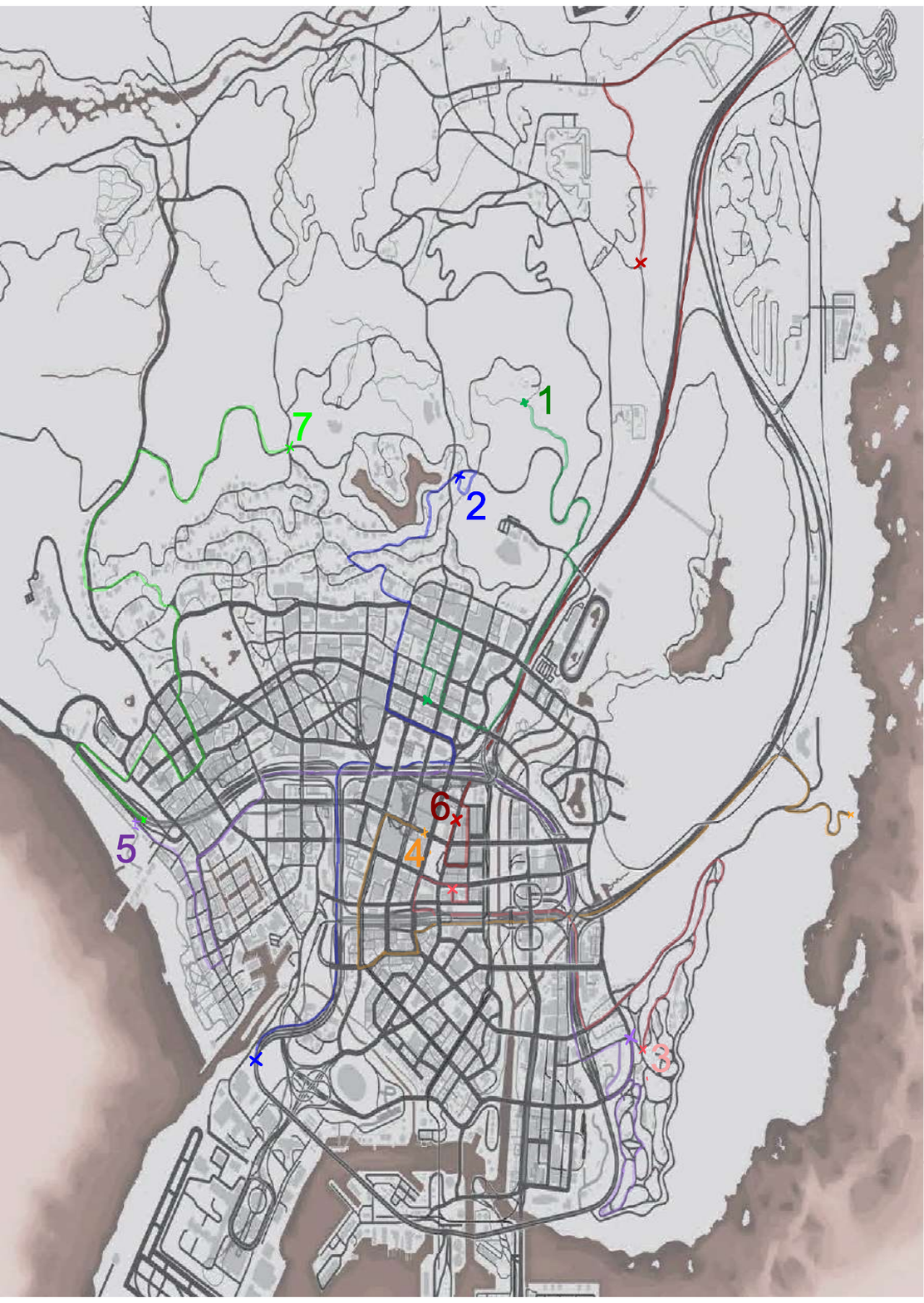}
 \caption{The routes of the GTAVs dataset, best seen in color.}
\label{fig:GTARoutes}
\end{figure}

Figure \ref{fig:GTAExamples} shows examples of the different classes.
The dataset contains seven recordings with dedicated routes that do not intersect each other to avoid overlaps between the training, validation and test split. Figure \ref{fig:GTARoutes} shows a map of the different routes.
All routes are recorded during the day hours and similar weather conditions.
Additionally, we designed the routes such that all classes are present in each route.
However, the share of classes in each route varies.
By sampling the routes at 10Hz, the dataset contains around 35000 frames. Each frame has a size of $128 \times 72$ pixels.
The dataset is split into an initial labeled pool, an unlabeled pool, and a validation and test set for the experiments.
The exact split is shown in Table \ref{tab:gtsvs}.

\begin{table}
\centering
 \begin{tabular}{m{0.25\textwidth} m{0.4\textwidth}}
  \multicolumn{1}{c}{Assignment} & \multicolumn{1}{c}{Sessions}\\
  \hline \\

    initial labeled pool & \text{Route6} \\
    unlabeled pool &  \text{Route2}, \text{Route4}, \text{Route5}, \text{Route7} \\
    validation set & \text{Route1}\\
    test set & \text{Route3}\\
 \end{tabular}
\caption{This table shows the dataset split into internal labeled and unlabeled pool training set as well as validation and test set for GTAVs. }
\label{tab:gtsvs}
\end{table}

\section{Experiment Details}
\label{ap:experiment}
In this section, we highlight our experiment details and hyperparameters. As the datasets are highly redundant due to the recording character, we applied early stopping on the validation accuracy, setting the parameter for patience to 30. We followed the official implementations of Resnet18  \citep{He2016} and VGG11 \citep{Simonyan15} provided by PyTorch \citep{pytorch}, except for minor modifications. For Resnet18, we added two fully connected layers with dropout layers between them to the head. For VGG11, we reduced the hidden layer size to 1024 and 512. Additionally,
the convolutional layers were initialized using the pre-trained ImageNet \citep{Deng2009} weights provided by PyTorch.
For the ResNet18 model, we attached the loss learning modules after each of the four blocks. In the case of VGG11, we attached the loss modules after the last four max pooling layers.

For GTAVs, we set the batch size to 128 and used a learning rate of 0.001 with SGD with a momentum of 0.9.
For A2D2s, we used a batch size of 64 with a reduced momentum of 0.8. As the redundancy of the dataset is quite high, we chose an early stopping strategy on the validation accuracy with patience 30. We used the checkpoint with the highest validation accuracy to continue. For the loss learning module, we split the early stopping to a first, detached the gradients from the loss module to the task model and a second early stopping to stop the training. Different detachment points are examined in Section \ref{ap:detachment}.
We set the margin $\zeta$ and $\xi$ to 0.5. The scale of the $\text{L1}_\text{margin}$ loss $\lambda$ is set to 0.5. The scaling factor $\eta$ is set to 1 for the combined loss.

Further, we neglected extensive augmentations for our active learning setup. As we are using a stream-based setup, using the normalization to zero mean and unit standard deviation of the whole training pool is not possible. As some changes between the drives can occur, we used histogram equalization as a data preprocessing step instead. Due to the high redundancy of the dataset, we only select between 15\% to 42\% of the training data. All experiments are conducted three times with the seeds 1,42,64. If not named differently, the parameters suggested by the authors are taken for the state-of-the-art methods. The number of forward passes for MC Dropout is set to 10.

The hyperparameters describe the weights in a combined loss function and are defined with a hyperparameter search. As the loss is a combined loss, the loss learning task influences the weights of the layers used for the perception task of the primary model. Therefore, they should be chosen such that strong regularizing and disturbance effects, leading to decreased task performance, are avoided. Too strong criteria on loss learning loss and a bad detachment point lead to underperformance in task results. The parameters can be appropriately selected by comparing a model with a loss module and a model without a loss module in the initial training. The second goal is tuning the parameters based on the ranking quality, whereas the first goal is more important, which can be achieved by a rough grid search. Our findings show that the approach is robust and not too sensitive to small changes.

 \section{Semantic Segmentation Experiments}
 \label{app:semseg}

This section highlights the exact settings of our semantic segmentation experiments based on object detection subset from A2D2 using a DeeplabV3 Chen \etal~\citep{Chen_2018_ECCV} model.
The loss learning layers are attached in the same way to the ResNet34 backbone as for ResNet18. Additionally, we added loss learning modules to each Atrous Spatial Pyramid Pooling (ASPP) block of the DeeplabV3 module.
We used a version of the dataset also containing the bounding box labels. The splits and query size are set identically to the parameter in the classification experiments. The same holds for TPL and loss learning parameters, which are taken out of the box from the classification experiments. For the backbone, pre-trained weights are used. For the training, we used a batch size of 32 and a learning rate of 0.05 for the head and 0.005 for the backbone. We did not use any augmentation but the same preprocessing as used in the classification experiments, except for the image size. The images have been resized to 640x400. We compared our TPL method to loss learning, a random selection and a MC dropout entropy-based selection using the mean of the pixel entropy values to determine the value of each sample. As the exact class remapping used in \cite{Geyer2020} is not given, we defined 17 classes and provided the mapping in Table \ref{tab:Class_Setting}. The classes "ignored" and "ego\_vehicle" are excluded from the evaluation.

\section{Ablation Study}
Additionally to the main experiments, we evaluated the training strategy in Section \ref{ap:ts}, the detachment of the loss learning module in Section \ref{ap:detachment} and our modifications to the loss learning module loss in Section \ref{ap:parameters}.

\begin{table}
\centering

 \begin{tabular}{ll}

Class Index & Classes \\
\hline
Ignored &

    Rain dirt,

    Blurred area \\

  Nature &

    Nature object  \\

  Buildings

          & Buildings\\

Traffic Guide&
Electronic traffic,
Irrelevant Signs,
Traffic guide obj. \\ &
Signal corpus,
Poles,
Grid structure \\
& Traffic signal 1
Traffic signal 2,
Traffic signal 3 \\&
Traffic sign 1,
Traffic sign 2,
Traffic sign 3
  \\
  Non Drivable & Non-drivable street, Parking area\\
  Ego Car &Ego car\\
  Bicycle &
    Bicycle 1,
    Bicycle 2,

    Bicycle 3,

    Bicycle 4

  \\

  Pedestrian &

    Pedestrian 1,

    Pedestrian 2,

    Pedestrian 3

  \\

  Small Moving Objects &

    Small vehicles 1,

    Small vehicles 2,

    Small vehicles 3

  \\

  Moving Medium Objects &

    Car 1,

    Car 2,

    Car 3,

    Car 4

  \\

  Moving Big Objects &

    Truck 1,

    Truck 2,

    Truck 3\\&

    Utility vehicle 1,

    Utility vehicle 2,

    Tractor

  \\

  Sky &
    Sky
  \\

  Street Areas &
   Speed Bumper,
    Driveable cobblestone \\&
    Slow drive area,
    RD normal street
  \\

  Guiding &
    Road blocks
  \\

  Lane Markings &
    Zebra crossing,
    RD restricted area,
    Painted driv. instr.
  \\

  Lines &
    Solid line,
    Dashed line
  \\

  Sidewalks &

    Sidebars,
    Curbstone,
    Sidewalk
  \\

  Obstacles &
    Obstacles / Trash,
    Animals
  \\
 \end{tabular}
 \caption{Class setting for semantic segmentation.}
 \label{tab:Class_Setting}
 \end{table}

\subsection{Training Strategy}
\label{ap:ts}
The topic of the training scheme of AL cycles is relatively unexplored. The model can be trained from scratch or the current model state can be enhanced and reused. As most works tend to retrain the model, we use this strategy. Additionally, we can avoid side effects due to continuous training strategies and focus on the selected properties in this way. As \cite{Schmidt2020} and \cite{Dayoub2017} reported interesting results by using different strategies, we conducted experiments to evaluate the decision of \citep{Yoo2019} to use a continuous training strategy. The results are reported in Table \ref{tab:ALstrategy}.

As can be seen, the continuous training strategy has a minor effect on all methods evaluated. While TPL and random are decreased at the last cycles and boosted at the first cycles, it is the other way around for loss learning. The inconsistency of the results shows how difficult an interpretation and a distinction between the selection method and training strategy is. Interestingly, the order of the methods does not change. Due to the small effect on performance, continuous training strategies are mainly suitable for time savings.
\begin{table}[b]
\centering

 \begin{tabular}{ccccc}

Method & Cycle 1& Cycle 2 &	Cycle 3 & Cycle 4 \\
  \hline \\
TPL retrain	& 0.802(8) & 0.923(3) & 0.962(5) & 0.967(2) \\
Loss learning retrain & 0.827(11) & 0.878(8) & 0.937(11) & 0.938(6) \\
Random retrain	& 0.839(18) & 0.918(5) & 0.950(9) &	0.956(5) \\
\hline \\
TPL continuous &	0.833(11) &	0.922(6) &	0.950(8) &	0.960(11) \\
Loss learning continuous &	0.813(17) &	0.879(19) &	0.938(12) &	0.940(3)\\
Random continuous & 0.844(4) & 0.902(14)	& 0.942(8)	& 0.949(6) \\

 \end{tabular}
 \caption{Different training strategies for active learning.}
 \label{tab:ALstrategy}
 \end{table}

\subsection{Loss Learning Module Detachment}
\label{ap:detachment}
\begin{figure}[hb!]
\centering
\subfigure[GTAVs ResNet18\label{fig:GTA_res18_detach_comp}]{%
   \includegraphics[width=0.32\textwidth]{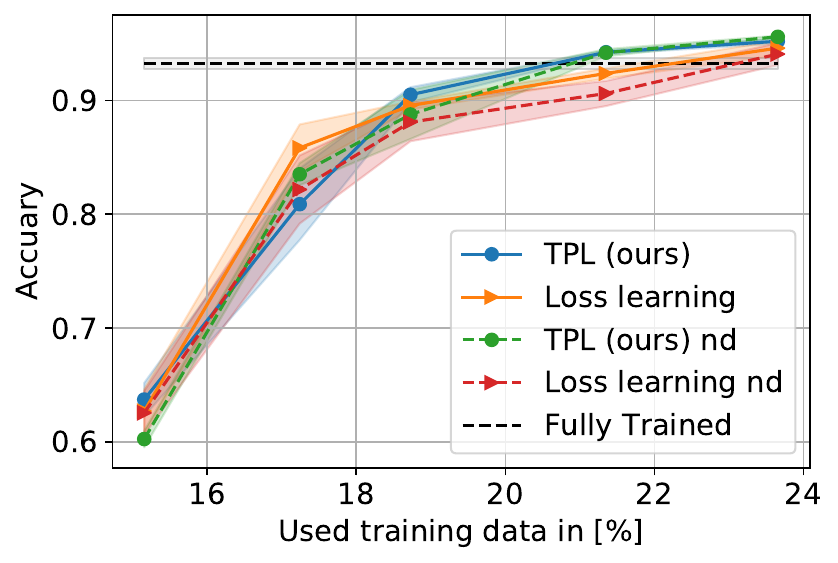}
 }%
 \subfigure[GTAVs VGG11\label{fig:GTA_vgg11_detach_comp}]{%
   \includegraphics[width=0.32\textwidth]{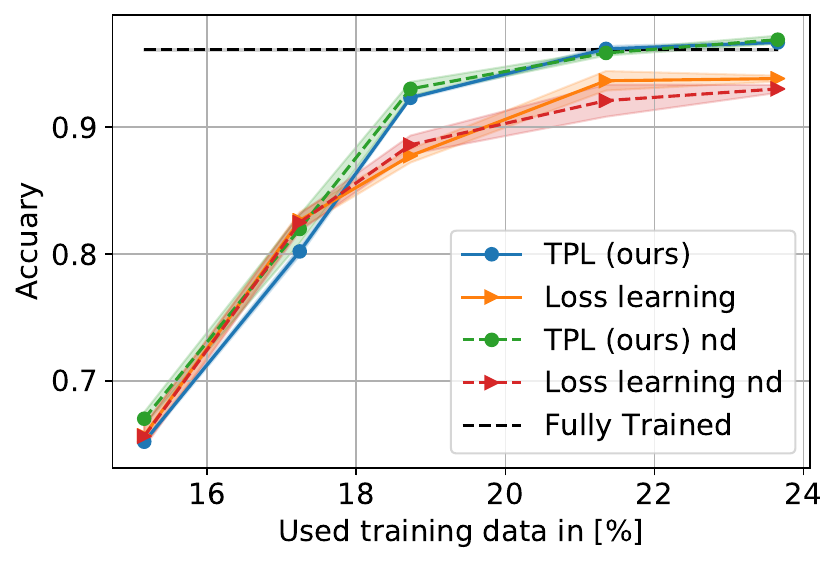}
}%
\subfigure[A2D2s VGG11 \label{fig:A2D2_vgg11_detach_comp}]{%
   \includegraphics[width=0.33\textwidth]{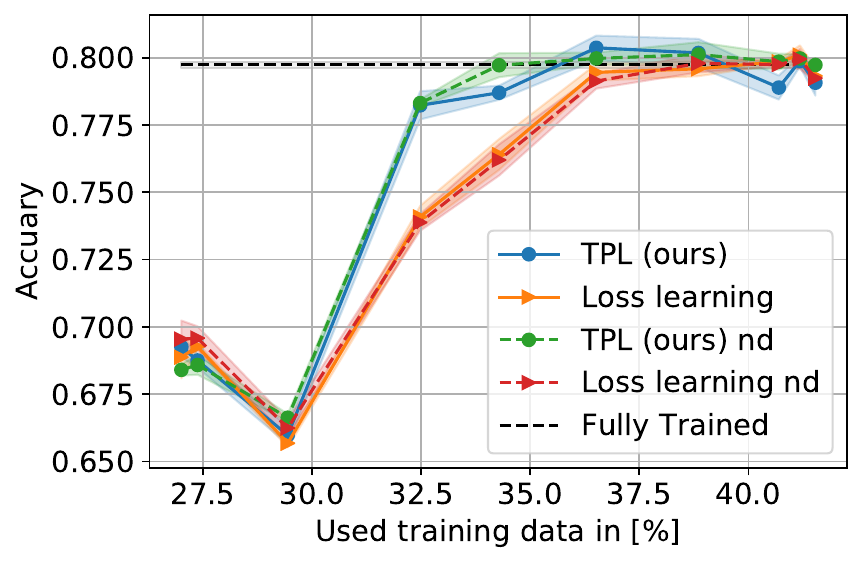}
 }
 \caption{Detachment comparison, non-detached is abbreviated with "nd".}
 \label{fig:detach_comp}
\end{figure}

In this section, we evaluate the point of detaching the gradients from the loss learning module. Fixed epochs, as proposed by \citep{Yoo2019}, are suboptimal for datasets with lower diversity as they tend to overfit. As described in Section \ref{ap:experiment}, we follow an early stopping strategy for the detachment.
To justify our proposed approach, we compared the early stopping detaching of gradients with the approach of not detaching at all and depicted the results in Figrue \ref{fig:detach_comp}.
The figures indicate that the performance of TPL is almost independent of the detachment point for the GTAVs dataset in the areas of higher-used training data. While at the first two phases, the accuracy of TPL differs. This can also be observed in the loss learning approach. However, the effect is more present for the region between the minimum and maximum data selection size. For A2D2s in Figure \ref{fig:A2D2_vgg11_detach_comp}, the effect on loss learning is neglectable, while TPL shows in the region of the fully trained network differences and achieves higher saving when not detached. For A2D2s, it should be considered that due to the independent recordings in the training, validation and test set, the distributions can have limitations in overlap. Therefore, the early stopping on the validation set has limitations.

Concluding the experiment shows that both methods are rather unsusceptible for the detachment point. It can be considered a hyperparameter for minor optimization purposes. The finding additionally underlines the downsides of learning-based approaches like VAAL \citep{Sinha2019}, loss learning \cite{Yoo2019} and CoreGCN \cite{Caramalau2021}, which showed increased data saving potential by the cost of additional models and hyperparameters.

\subsection{Loss Learning Loss Modifications}
\label{ap:parameters}
In this section, we evaluate our modification to the loss learning loss and highlight the influence of different parameters.

\textbf{L1 regularization:}

\begin{figure}[hbt]
\centering
\subfigure[Loss function modification with vanilla comparison for ResNet18 on GTAVS.\label{fig.lossFuncAbl}]{%
   \includegraphics[width=0.33\textwidth]{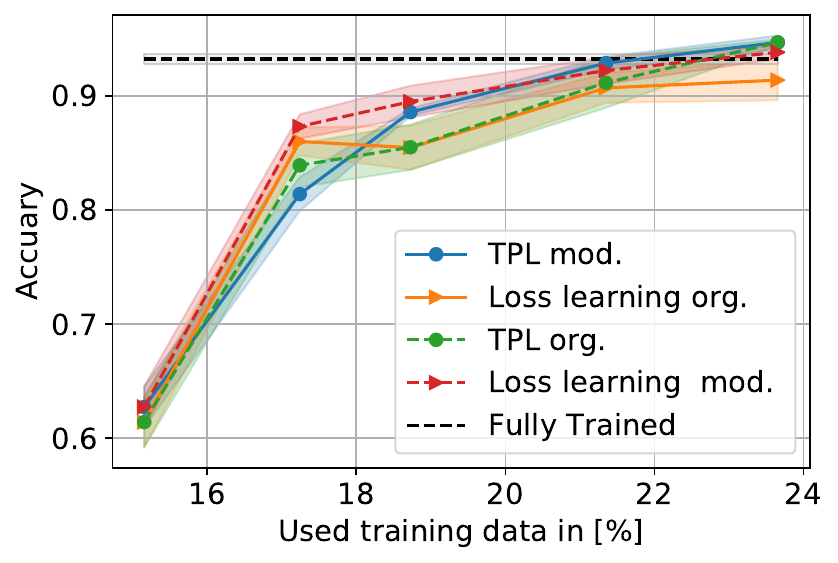}
 }%
 \subfigure[Loss function margin values $\xi$ and $\zeta$ study for ResNet18 on GTAVS.\label{fig.margin}]{%
   \includegraphics[width=0.33\textwidth]{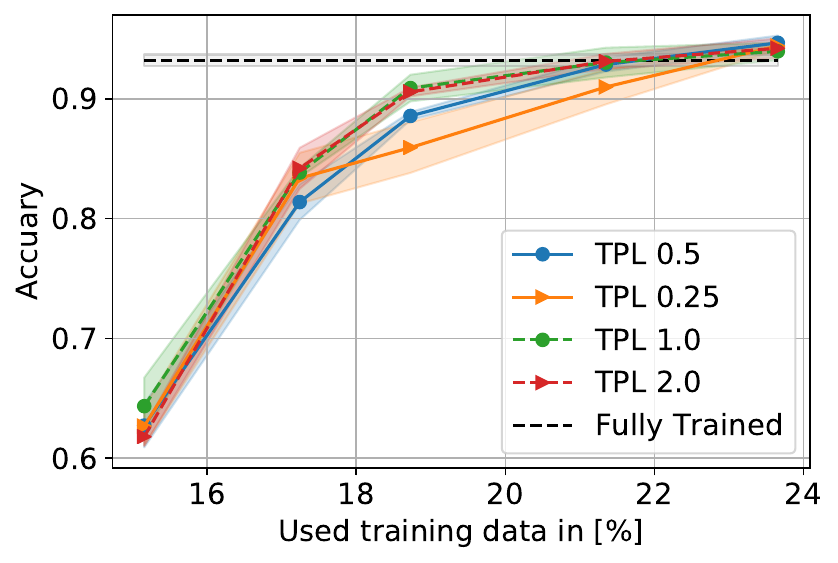}
}%
\subfigure[Lambda factor comparison for ResNet18 on GTAVS. \label{fig:lambda}]{%
   \includegraphics[width=0.33\textwidth]{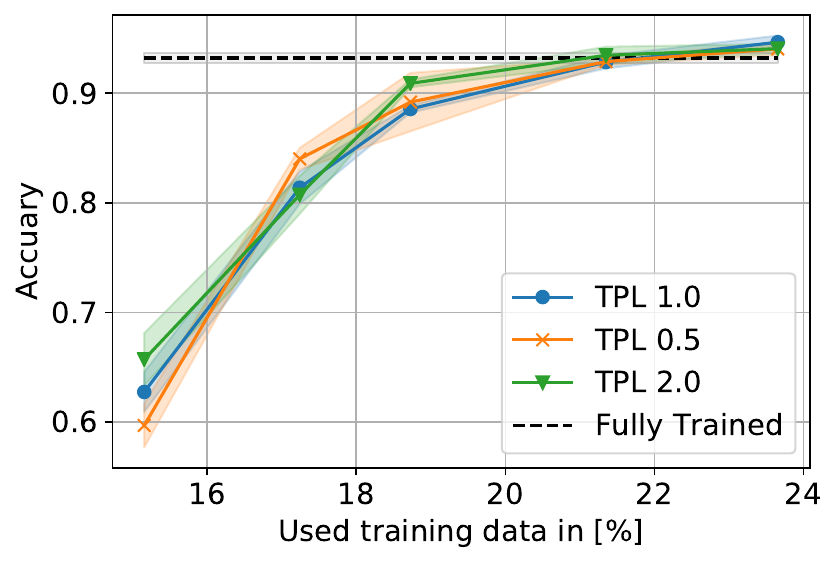}
 }
 \caption{Ablation study of different parameters of the loss learning loss function. }
 \label{fig:parameter_comp}
\end{figure}

In order to evaluate the effect of our L1 regularization adaption to the loss learning module loss function in Equation \ref{eq:loss_module}, we compare TPL and the vanilla loss learning with both losses in Figure \ref{fig.lossFuncAbl}. The plot shows that our loss function adaption, as well as our Temporal predictive loss function, both improve the vanilla approach individually. As the combined approach delivers the best performance, both effects are cumulative.

\textbf{Margin selection:}

In the following study, we highlight the effect of the Xi and Zeta margin in Equation \ref{eq:loss_module}. For simplification, we use for $\zeta$ and $\xi$ the same value and show their effects in Figure \ref{fig.margin}. As the figure indicates, the margin factors have a minor role compared to the loss function. While higher margins improve the performance in low-data regions, our selected value shows the highest accuracy. A more fine-grained parameter search would potentially improve the results further. Conversely, the limited effect of tuning this parameter indicates that a low-effort parameter search is sufficient, making our method easier to apply.

\textbf{Lamda factor selection:}

Lastly, we examine the lambda factor for the regularization of the loss learning module loss. Figure \ref{fig:lambda} shows different values for this factor and indicates that the lambda factor has an
impact on regions of low data, which is reduced for regions with more data. While higher regularization improves the performance in low-data regions, lower and higher regularization improves the result for low-data regions. The differences between different lambda values are strongly decreasing for regions with more data. We choose based on the highest accuracy the value one, which has also been chosen by Yoo~\etal \cite{Yoo2019}. Also, in this case, a more fine-grained parameter search has the potential to improve the results further. However, the limited effect of tuning this parameter shows that a low-effort parameter search is sufficient, proving again that our method is easy to apply.

\end{document}